\documentclass[10pt,twocolumn,letterpaper]{article}

\usepackage{iccv}
\usepackage[accsupp]{axessibility}
\usepackage{times}
\usepackage{epsfig}
\usepackage{graphicx}
\usepackage{amsmath}
\usepackage{amssymb}

\usepackage{multicol}
\usepackage{multirow}
\usepackage{booktabs}
\usepackage{utfsym}
\usepackage{makecell}
\usepackage{subcaption}
\usepackage{colortbl}
\usepackage{appendix}

\usepackage{tikz,pgfplots,pgfplotstable}
\usepackage{color}
\usepgfplotslibrary[groupplots]
\definecolor{mediumelectricblue}{rgb}{0.12,0.314,0.588}
\definecolor{mossgreen2}{RGB}{138,154,91}
\definecolor{internationalorange}{RGB}{100,31,0}
\definecolor{lightsalmon}{RGB}{255,160,122}

\definecolor{selfyellow}{RGB}{255,209,0}
\definecolor{selfblue}{RGB}{89,138,234}
\definecolor{selfgreen}{RGB}{133,235,133}

\usepackage[pagebackref=true,breaklinks=true,letterpaper=true,colorlinks,bookmarks=false]{hyperref}

\iccvfinalcopy % *** Uncomment this line for the final submission

\def\iccvPaperID{2506} % *** Enter the ICCV Paper ID here

\ificcvfinal\fi

\newlength\savewidth\newcommand\shline{\noalign{\global\savewidth\arrayrulewidth
  \global\arrayrulewidth 1pt}\hline\noalign{\global\arrayrulewidth\savewidth}}

\begin{document}

%%%%%%%%% TITLE
\title{Group Pose: A Simple Baseline for End-to-End Multi-person Pose Estimation}
\newcommand*\samethanks[1][\value{footnote}]{\footnotemark[#1]}
% \thanks{Corresponding authors.} \samethanks 
\author{
Huan Liu$^{1,3}$\thanks{Equal Contribution. Work done when H. Liu is an intern at Baidu.}\quad
Qiang Chen$^{2}$\samethanks\quad
Zichang Tan$^{2}$\quad
Jiang-Jiang Liu$^{2}$\quad
Jian Wang$^{2}$\quad
Xiangbo Su$^{2}$\\
Xiaolong Li$^{1,3}$\quad
Kun Yao$^{2}$\quad
Junyu Han$^{2}$\quad
Errui Ding$^{2}$\quad
Yao Zhao$^{1,3}$\thanks{Corresponding author. Email: yzhao@bjtu.edu.cn}\quad
Jingdong Wang$^2$\\[1.2mm]
$^1$Institute of Information Science, Beijing Jiaotong University \quad
$^2$Baidu VIS \\
$^3$Beijing Key Laboratory of Advanced Information Science and Network Technology, Beijing, China\\
}

\maketitle
\ificcvfinal\thispagestyle{empty}\fi

\begin{abstract}
In this paper, we study the problem of end-to-end multi-person pose estimation. State-of-the-art solutions adopt the DETR-like framework, and mainly develop the complex decoder, {\em e.g.}, regarding pose estimation as keypoint box detection and combining with human detection in ED-Pose~\cite{yang2023explicit}, hierarchically predicting with pose decoder and joint (keypoint) decoder in PETR~\cite{shi2022end}.

We present a simple yet effective transformer approach, named Group Pose. We simply regard $K$-keypoint pose estimation as predicting a set of $N\times K$ keypoint positions, each from a keypoint query, as well as representing each pose with an instance query for scoring $N$ pose predictions.

Motivated by the intuition that the interaction, among across-instance queries of different types, is not directly helpful, we make a simple modification to decoder self-attention. We replace single self-attention over all the $N\times(K+1)$ queries with two subsequent group self-attentions: (i) $N$ within-instance self-attention, with each over $K$ keypoint queries and one instance query, and (ii) $(K+1)$ same-type across-instance self-attention, each over $N$ queries of the same type. The resulting decoder removes the interaction among across-instance type-different queries, easing the optimization and thus improving the performance. Experimental results on MS COCO and CrowdPose show that our approach without human box supervision is superior to previous methods with complex decoders, and even is slightly better than ED-Pose that uses human box supervision. Code is available here\footnote{Code of \href{https://github.com/Michel-liu/GroupPose-Paddle}{Paddle} and \href{https://github.com/Michel-liu/GroupPose}{PyTorch} implementations.}.
\end{abstract}

\section{Introduction} \label{sec:intro}
Multi-person pose estimation aims to detect the corresponding human keypoints for all human instances in an image. Previous frameworks include top-down~\cite{newell2016stacked,fang2017rmpe,xiao2018simple,chen2018cascaded,sun2019deep,wang2020deep} and bottom-up methods~\cite{cao2017realtime,newell2017associative,kreiss2019pifpaf,cheng2020higherhrnet} that divide the task into two sequential sub-tasks: human detection with single-person pose estimation or human-agnostic keypoint detection with human instance grouping~\cite{newell2017associative}. Another line in previous frameworks is one-stage methods~\cite{mao2021fcpose,nie2019single,tian2019directpose,wei2020point}, which directly predict instance-aware keypoints. These frameworks rely on non-differentiable hand-crafted post-processes~\cite{hosang2017learning,newell2017associative}, which complicate the pipelines and challenge the optimizations. Inspired by the success of DETR~\cite{carion2020end} in object detection, building an end-to-end framework for multi-person pose estimation has seen significant interest.

\begin{figure}[t]
\centering
\includegraphics[width=\linewidth]{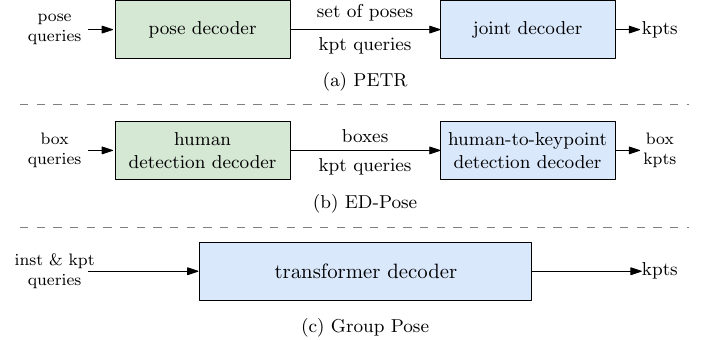}
\captionsetup{font=small}
\caption{\textbf{Comparison of transformer decoders.} Here we mainly illustrate the overview of the decoder part of PETR~\cite{shi2022end}, ED-Pose~\cite{yang2023explicit}, and our Group Pose. The three end-to-end frameworks differ in query design and decoder architecture. Group Pose only uses a simple transformer decoder rather than developing complex decoders. `inst' and `kpt' represent for instance and keypoint.}
\label{fig:figure1}
\vspace{-3mm}
\end{figure}

Recent approaches 
follow the DETR framework~\cite{carion2020end,zhu2020deformable,zhang2022dino},
with the transformer encoder-decoder architecture
for multi-person pose estimation, as shown in Figure~\ref{fig:figure1}.
PETR~\cite{shi2022end} hierarchically 
predicts the keypoint positions
and uses two subsequent decoders, pose decoder
and joint decoder,
with two different queries,
pose query (a person has one pose query) for pose decoder,
and keypoint queries for joint decoder.
ED-Pose~\cite{yang2023explicit} 
transfers pose estimation 
to a keypoint box detection problem,
and learns a content query and a box query for each keypoint position prediction
with using the box size 
to process the query.

% overview of detr-based decoder

In this paper, we 
present a simple yet effective transformer approach,
named Group Pose,
for end-to-end multi-person pose estimation.
Instead of using a single query
to predict and score one pose, 
similar to ED-Pose,
we use the $N\times K$ keypoint queries
to regress the $N\times K$ positions,
by regarding each keypoint as an object,
as well as 
$N$ instance queries,
each representing 
a $K$-keypoint pose for 
scoring the $K$-keypoint pose prediction.

We make a simple modification 
for the decoder architecture. 
We replace standard self-attention in the decoder
with two subsequent group self-attentions:
$N$ parallel self-attentions
with each over $K$ keypoint queries
and the corresponding instance query
for exploiting kinematic relation and gathering information for scoring pose predictions,
and $(K+1)$ parallel self-attentions
with each over $N$ queries of the same type
for collecting duplicate prediction information like self-attention
of the original DETR.

The two self-attentions capture
two kinds of interactions:
(i) $N$ within-instance interactions
over $K$ keypoint queries
and one instance query,
(ii) $(K+1)$ across-instance interactions
over $N$ queries of the same type
({\em e.g.}, nose keypoint query or instance query).
The extra interactions in standard self-attention
are (iii) across-instance interactions
for queries 
with different types,
which is not directly useful.
Empirical results show that
the removal of the third kind of interactions
eases the optimization and thus 
improves the performance.

The design about self-attention
is different from
the closely-related approach ED-Pose~\cite{yang2023explicit}.
On the one hand,
in addition to removing the third interactions,
ED-Pose only models 
across-instance interactions 
for the instance queries.
In contrast, our approach also models 
across-instance interactions
for the $K$ keypoint types.
On the other hand,
our approach separates the two interactions
using two subsequent group self-attentions,
explicitly exploring the information
about queries belonging to the same human instance, 
and with the same type.
ED-Pose couples the two interactions
using a single masked self-attention
with the third interactions masked.
It is empirically demonstrated that
the two differences benefit the pose estimation performance.

Experimental results show that our simple approach Group Pose without human box supervision surprisingly outperforms the recent end-to-end methods with human box supervisions on MS COCO~\cite{lin2014microsoft} and CrowdPose~\cite{li2019crowdpose}. Notably, Group Pose achieves $72.0$ AP with ResNet-50~\cite{he2016deep} and $74.8$ AP with Swin-Large~\cite{liu2021swin} on MS COCO \texttt{val2017}. We hope our simple transformer decoder in Group Pose will motivate people to simplify the design in end-to-end multi-person pose estimation.

\section{Related Work}
Multi-person pose estimation is a challenging task that aims to detect the corresponding human keypoints for all human instances in an image. Previous methods usually adopt complex frameworks to address it, which are divided into non-end-to-end and end-to-end methods.

\vspace{1mm}
\noindent\textbf{Non-end-to-end methods.} There are typically two types: two-stage and one-stage. Two-stage frameworks, including top-down~\cite{newell2016stacked,fang2017rmpe,xiao2018simple,chen2018cascaded,sun2019deep,wang2020deep} and bottom-up methods~\cite{cao2017realtime,newell2017associative,kreiss2019pifpaf,cheng2020higherhrnet}, split the multi-person pose estimation task into two sequential sub-tasks, human detection with single-person pose estimation or human-agnostic keypoint detection with human instance grouping. In top-down methods, an object detector is first employed to detect the boxes of human instances, which is then cropped for single-person pose estimation in each box. Bottom-up methods first predict all human keypoints in a human-agnostic way then group them into instances. While one-stage frameworks~\cite{zhou2019objects,mao2021fcpose,nie2019single,tian2019directpose,wei2020point} directly predict instance-aware keypoints. These methods require hand-crafted pose-processes, such as NMS~\cite{hosang2017learning} or grouping~\cite{newell2017associative}, which complicate the pipelines and challenge the optimizations. In this paper, we focus on concise ways, end-to-end frameworks.

\vspace{1mm}
\noindent\textbf{End-to-end methods.} Current end-to-end multi-person pose estimation frameworks~\cite{shi2022end,xiao2022querypose,yang2023explicit} are built by following the designs of DETR~\cite{carion2020end} and its variants~\cite{zhu2020deformable,meng2021conditional,chen2022group,liu2022dab,li2022dn,zhang2022dino}. They adopt a paradigm of splitting the multi-person pose estimation task into two sub-processes. For example, PETR~\cite{shi2022end} views the task as a hierarchical set prediction problem. It first determines human instances by predicting a set of poses with a pose decoder and then refines the keypoints in each pose with a joint (keypoint) decoder. There are also two different types of queries, pose query (a person has one pose query) for pose decoder and keypoint queries for joint decoder. QueryPose~\cite{xiao2022querypose} and  ED-Pose~\cite{yang2023explicit} follow this end-to-end paradigm but further incorporate an extra human detection task. QueryPose~\cite{xiao2022querypose} follows Sparse R-CNN~\cite{sun2021sparse} to build two parallel RoIAlign-based~\cite{he2017mask} decoders to perform human detection and pose estimation, respectively. ED-Pose \cite{yang2023explicit} transfers pose estimation task to a keypoint box detection problem. It first employs a human detection decoder~\cite{zhang2022dino} to determine the human instances with box queries. Then it builds a human-to-keypoint detection decoder with a content query and a box query for each keypoint position, collecting contextual information near keypoint positions.

Although these end-to-end methods show promising results in multi-person pose estimation, they rely on complex decoders. Our Group Pose, on the other hand, adopts a simple transformer decoder, improving the performance and simplifying the process.

% overview of group pose
\begin{figure*}[t]
\centering
\includegraphics[width=0.9\textwidth]{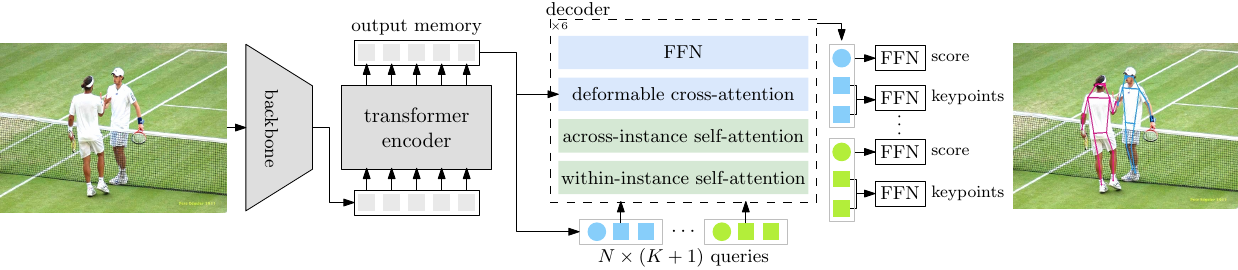}
\captionsetup{font=small}
\caption{\textbf{Our Group Pose architecture.} The backbone takes an image as input and outputs image features, which are refined by the transformer encoder. To directly predict $N$ human poses with $K$ keypoint positions in each pose, we adopt $N\times (K+1)$ queries, containing $N$ instance queries for scoring poses and $N\times K$ keypoint queries for regressing positions.}
\label{fig:overview}
\vspace{-5mm}
\end{figure*}

\section{Group Pose}
Group Pose is a simple yet effective end-to-end multi-person pose estimation framework. We follow previous end-to-end frameworks~\cite{shi2022end,yang2023explicit} to view the multi-person pose estimation task as a set prediction problem, but directly adopt a simple transformer decoder~\cite{zhu2020deformable} instead of complex decoders, simplifying the process. Next, we introduce the key elements of Group Pose.

\subsection{Overview}
The overall structure of Group Pose is depicted in Figure~\ref{fig:overview}. Group Pose consists of a backbone~\cite{he2016deep,liu2021swin}, a transformer encoder~\cite{vaswani2017attention}, a transformer decoder, and task-specific prediction heads. This framework enables Group Pose to simultaneously regresses $K$ keypoints ({\em e.g.}, $K=17$ on MS COCO) for $N$ human instances given an image.

\vspace{1mm}
\noindent \textbf{Backbone and transformer encoder.} We directly follow DETR frameworks~\cite{zhu2020deformable} to build the backbone and the transformer encoder (with 6 deformable transformer layers~\cite{zhu2020deformable}) for Group Pose. It takes an image as input and outputs the extracted multi-level features, which serve as inputs for the following transformer decoder. We use 4 feature levels in Group Pose, with downsampling rates of $\{8, 16, 32, 64\}$.

\vspace{1mm}
\noindent \textbf{Transformer decoder.} For transformer decoder, we adopt a combination of $N\times K$ keypoint queries and $N$ instance queries as input instead of using a single query to predict and score one pose. The keypoint queries regard each keypoint as an object and are used to regress the $N\times K$ keypoint positions, while each instance query is for scoring the corresponding $K$-keypoint pose prediction. Besides, the architecture of transformer decoder is simple, which stacks $6$ same decoder layers~\cite{zhu2020deformable}. In each decoder layer, we follow the macro design of previous DETR frameworks by building self-attention, cross-attention implemented with deformable attention, and FFN. We only make simple modifications to self-attention in our Group Pose. Specifically, we replace the standard self-attention with two subsequent group self-attentions, enabling decoder layers to perform interactions over queries belonging to the same human instance and with the same type.

\vspace{1mm}
\noindent \textbf{Prediction heads.} There are two prediction heads implemented with FFNs in Group Pose for human classification and human keypoints regression. Group Pose predicts for $N$ human poses, each contains a classification score and $K$ keypoint positions for the corresponding $K$-keypoint pose.

\vspace{1mm}
\noindent \textbf{Loss function.} The Hungarian matching algorithm \cite{carion2020end} is employed for one-to-one assignment between predicted poses and ground-truth poses. Our loss function comprises solely of classification loss ($\mathcal{L}_{cls}$) and keypoint regression loss ($\mathcal{L}_{kpt}$), without any extra supervisions such as human detection loss in QueryPose~\cite{xiao2022querypose} and ED-Pose~\cite{yang2023explicit} or heatmap loss in PETR~\cite{shi2022end}. The keypoint regression loss ($\mathcal{L}_{kpt}$) is a combination of a normal $\ell_1$ loss and a constrained $\ell_1$ loss named Object Keypoint Similarity (OKS)~\cite{shi2022end}. We directly use the cost coefficients and loss weights of ED-Pose~\cite{yang2023explicit} in the hungarian matching and the calculation of losses.

\subsection{$N\times K$ keypoint queries and $N$ instance queries} \label{subsec:queries}
In multi-person pose estimation, frameworks are required to predict $N$ human poses with $K$ keypoint positions in each pose given an image. We directly use $N\times K$ keypoints queries to predict the poses, with each $K$ predicted keypoint positions to represent a corresponding $K$-keypoint pose for a human instance. To classify if it is a human instance by scoring the predicted $N$ human poses, we also introduce $N$ instance queries. Thus, in Group Pose, a human instance can be represented with a combination of $K$ keypoint queries and one instance query. As these two types of queries are responsible for different tasks, we construct and initialize them differently.
% overview pf group pose
\begin{figure*}[t]
\centering
\includegraphics[width=0.8\textwidth]{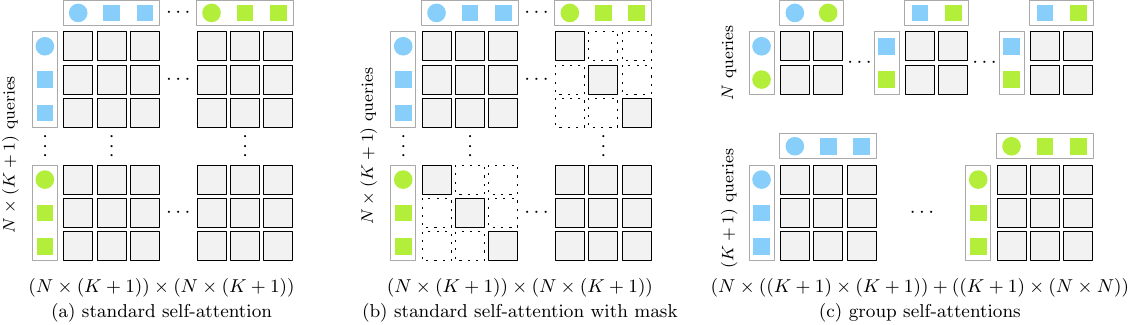}
\captionsetup{font=small}
\caption{\textbf{Conceptual comparison of self-attention implementations.} Given $N \times (K + 1)$ queries, three different implementations of self-attention are considered in the decoder layers. (a) standard self-attention over all queries. (b) standard self-attention with an attention mask, removing the across-instance interactions for queries with different types. (c) our proposed group self-attentions. Dashed boxes indicate the masked-out interactions.}
\label{fig:interaction}
\vspace{-5mm}
\end{figure*}

\vspace{1mm}
\noindent \textbf{Query construction.} We follow the previous end-to-end frameworks~\cite{shi2022end,yang2023explicit} to first identify human instances and predict human poses in each position of the output memory from the transformer encoder. Then we select $N$ human instances ($N=100$) based on the classification scores, resulting $N$ human poses and the corresponding output memory features of the selected $N$ positions (with the shape of $N\times D$, where $D$ is the channel dimension). We then construct and initialize the keypoint queries and the instance queries based on the above results. 

For $N\times K$ keypoint queries, we construct the content part of $K$ keypoint queries ($K\times D$) in each human instance by combining $K$ randomly initialized learnable keypoint embeddings ($K\times D$) and the corresponding output memory feature ($1\times D$). And the position part of $K$ keypoint queries ($K\times 2$) in each human instance is initialized with the corresponding predicted $K$-keypoint pose ($K\times 2$). For $N$ instance queries, we only consider the content part and use a randomly initialized learnable instance embedding ($1\times D$) for each human instance. This is because instance query is for the classification task, which does not requires explicit position information. When performing cross-attention in the decoder layers, we simply use the mean of $K$ keypoint positions as the reference point for the instance query in each human instance.

\subsection{Group self-attentions}
In DETR frameworks~\cite{carion2020end,zhu2020deformable,meng2021conditional}, self-attention in the transformer decoder is usually applied to model interactions among queries, collecting information from other queries and facilitating the duplicate removal for instances. While the situation is slightly different in Group Pose, as there two different types of queries, $K$ keypoint queries and one instance query for each human instance. Motivated by the intuition that the interactions among
across-instance queries of different types, {\em e.g.}, the interactions between the instance query and the keypoint queries across human instances, may not be directly helpful for the above purposes. We thus replace the standard self-attention over all the $N\times (K+1)$ queries with two subsequent group self-attentions: $N$ within-instance self-attentions and $K+1$ same-type across-instance self-attentions.

\vspace{1mm}
\noindent \textbf{$N$ within-instance self-attentions.} We build interactions among the queries within a human instance, exploiting kinematic relations and gathering information for scoring pose predictions. It calculates $N$ self-attention maps for $N$ human instances in parallel with shapes of $(K+1) \times (K+1)$, omitting the dimensions of batch size and attention heads.

\vspace{1mm}
\noindent \textbf{$(K+1)$ same-type across-instance self-attentions.} Similar to the within-instance group self-attentions, We add another group self-attentions, which collect information from the same-type queries in other instances and help remove duplicate predictions. We build interactions across human instances with the same-type queries, the instance query and each keypoint queries, resulting in $K+1$ same-type across-instance self-attentions. It also can be implemented in parallel with one self-attention module, calculating $K+1$ self-attention maps with shapes of $N\times N$.

Compared with the standard self-attention, our two subsequent group self-attentions explicitly explores the information about queries belong to the same human instance and with the same type, while remove the across-instance interactions for queries with different types. Empirical results in Section~\ref{sec:ablation} show that the removal of this kind of interactions eases the optimization and thus improves the performance for our Group Pose.

\section{Experiments}
%-------------------------------------------------------------------------
% main tables on COCO val
\begin{table*}[t]
\centering
\setlength{\tabcolsep}{9.5pt}
\renewcommand{\arraystretch}{1.25}
\footnotesize
\captionsetup{font=small}
\caption{\textbf{Comparisons with state-of-the-art methods on MS COCO \texttt{val2017}.} We also provide the reference (Ref) for previous frameworks. The `HM', `BR', and `KR' denote heatmap-based losses, human box regression losses, and keypoint regression losses. `RLE' represents the residual log-likelihood estimation in Poseur~\cite{mao2022poseur}. $\dagger$ denotes the flipping test.  $\ddagger$ removes the prediction uncertainty estimation in Poseur as a fair regression comparison. The \textbf{best results} are highlighted in \textbf{bold}.}
\begin{tabular}{c|c|l|c|l|l|ccc|cc}
\multicolumn{3}{c|}{Method} & Ref & Backbone & Loss & AP & AP$_{50}$ & AP$_{75}$ & AP$_{M}$ & AP$_{L}$ \\
\shline
\multirow{12}{*}{\rotatebox{90}{Non-End-to-End}} & \multirow{5}{*}{\rotatebox{90}{Top-Down}} & Mask R-CNN \cite{he2017mask} & CVPR 17 & ResNet-50 & HM  & $ 65.5 $ & $ 87.2 $ & $ 71.1 $ & $ 61.3 $ & $ 73.4 $ \\
& & Mask R-CNN \cite{he2017mask} & CVPR 17 & ResNet-101 & HM  & $ 66.1 $ & $ 87.4 $ & $ 72.0 $ & $ 61.5 $ & $ 74.4 $ \\
& & PRTR$^\dagger$ \cite{li2021pose} & CVPR 21 & ResNet-50 & KR  & $ 68.2 $ & $ 88.2 $ & $ 75.2 $ & $ 63.2 $ & $ 76.2 $ \\
& & Poseur$^\ddagger$ \cite{mao2022poseur} & ECCV 22 & ResNet-50 & RLE & $ 70.0 $ & $ - $ & $ - $ & $ - $ & $ - $ \\
& & Poseur \cite{mao2022poseur} & ECCV 22 & ResNet-50 & RLE & $ 74.2 $ & $ 89.8 $ & $ 81.3 $ & $ 71.1 $ & $ 80.1 $ \\
\cline{2-11}
& \multirow{4}{*}{\rotatebox{90}{Bottom-Up}} & HrHRNet$^\dagger$ \cite{cheng2020higherhrnet} & CVPR 20 & HRNet-w32 & HM & $ 67.1 $ & $ 86.2 $ & $ 73.0 $ & $ 61.5 $ & $ 76.1 $ \\
& & DEKR$^\dagger$ \cite{geng2021bottom}      & CVPR 21 & HRNet-w32 & HM & $ 68.0 $ & $ 86.7 $ & $ 74.5 $ & $ 62.1 $ & $ 77.7 $ \\
& & SWAHR$^\dagger$ \cite{luo2021rethinking}  & CVPR 21 & HRNet-w32 & HM & $ 68.9 $ & $ 87.8 $ & $ 74.9 $ & $ 63.0 $ & $ 77.4 $ \\
& & LOGO-CAP$^\dagger$ \cite{xue2022learning} & CVPR 22 & HRNet-w32 & HM & $ 69.6 $ & $ 87.5 $ & $ 75.9 $ & $ 64.1 $ & $ 78.0 $ \\
\cline{2-11}
& \multirow{4}{*}{\rotatebox{90}{One-Stage}} & DirectPose \cite{tian2019directpose} & $-$ & ResNet-50 & KR & $ 63.1 $ & $ 85.6 $ & $ 68.8 $ & $ 57.7 $ & $ 71.3 $ \\
& & CenterNet$^\dagger$ \cite{zhou2019objects} & $-$ & Hourglass-104 & KR+HM & $ 64.0 $ & $ - $ & $ - $ & $ - $ & $ - $ \\
& & FCPose \cite{mao2021fcpose} & CVPR 21 & ResNet-50 & KR+HM & $ 63.0 $ & $ 85.9 $ & $ 68.9 $ & $ 59.1 $ & $ 70.3 $ \\
& & InsPose \cite{shi2021inspose} & ACM MM 21 & ResNet-50 & KR+HM & $ 63.1 $ & $ 86.2 $ & $ 68.5 $ & $ 58.5 $ & $ 70.1 $ \\
\shline
\multirow{9}{*}{\rotatebox{90}{End-to-End}} & \multirow{6}{*}{\rotatebox{90}{Previous Works}} & PETR \cite{shi2022end} & CVPR 22  & ResNet-50 & HM+KR & $ 68.8 $ & $ 87.5 $ & $ 76.3 $ & $ 62.7 $ & $ 77.7 $ \\ 
& & PETR \cite{shi2022end} & CVPR 22 & Swin-L    & HM+KR & $ 73.1 $ & $ 90.7 $ & $ 80.9 $ & $ 67.2 $ & $ 81.7 $ \\
& & QueryPose \cite{xiao2022querypose} & NeurIPS 22 & ResNet-50 & BR+RLE & $ 68.7 $ & $ 88.6 $ & $ 74.4 $ & $ 63.8 $ & $ 76.5 $ \\
& & QueryPose \cite{xiao2022querypose} & NeurIPS 22 & Swin-L    & BR+RLE & $ 73.3 $ & $ 91.3 $ & $ 79.5 $ & $ 68.5 $ & $ 81.2 $ \\
& & ED-Pose \cite{yang2023explicit} & ICLR 23 & ResNet-50 & BR+KR & $ 71.6 $ & $ 89.6 $ & $ 78.1 $ & $ 65.9 $ & $ 79.8 $ \\
& & ED-Pose \cite{yang2023explicit} & ICLR 23 & Swin-L    & BR+KR & $ 74.3 $ & $ 91.5 $ & $ 81.6 $ & $ 68.6 $ & $ 82.6 $ \\
\cline{2-11}
& \multirow{3}{*}{\rotatebox{90}{Ours}} & GroupPose   & $-$     & ResNet-50 & KR     & $ 72.0 $ & $ 89.4 $ & $ 79.1 $ & $ 66.8 $ & $ 79.7 $ \\
& & GroupPose   & $-$     & Swin-T    & KR     & $ 73.6 $ & $ 90.4 $ & $ 80.5 $ & $ 68.7 $ & $ 81.2 $ \\
& & GroupPose   & $-$     & Swin-L    & KR     & $\bf 74.8 $ & $\bf 91.6 $ & $\bf 82.1 $ & $\bf 69.4 $ & $\bf 83.0 $ \\
\end{tabular}
\label{tab:table1}
\vspace{-3mm}
\end{table*}
%-------------------------------------------------------------------------
\subsection{Settings}
\noindent \textbf{Datasets.} Our experiments are conducted on two representative human pose estimation datasets, MS COCO~\cite{lin2014microsoft} and CrowdPose~\cite{li2019crowdpose}. MS COCO contains $200$K images and $250$K person instances with $17$ keypoint annotations per instance (we set the number of keypoint queries as $K=17$ on MS COCO). CrowdPose has $20$K images and $80$K person instances with $14$ keypoint annotations per instance ($K=14$ on CrowdPose). CrowdPose is more challenging as it includes many crowd and occlusion scenes. We train Group Pose on COCO \texttt{train2017} set and evaluate it on COCO \texttt{val2017} set and \texttt{test-dev} set. On CrowdPose, we train our model on the \texttt{train} set and evaluate it on the \texttt{test} set.

\vspace{1mm}
\noindent \textbf{Evaluation metric.} The OKS-based average precision (AP) scores are reported as the main metric for both datasets. For MS COCO, we adopt AP with different thresholds and different object sizes (medium and large), denoted as AP, AP$_{50}$, AP$_{75}$, AP$_{M}$, and AP$_{L}$, following the standard evaluation process\footnote{https://cocodataset.org/\#keypoints-eval}. On CrowdPose, to better evaluate the model performance in different
crowded scenarios, we adopt AP with different thresholds and different crowding levels, denoted as AP, AP$_{50}$, and AP$_{75}$, as well as AP$_{E}$, AP$_{M}$ and AP$_{H}$ for images with easy, medium and hard crowding levels.

\vspace{1mm}
\noindent \textbf{Implementation details.} Our training and testing settings follow ED-Pose~\cite{yang2023explicit}. During training, we adopt the widely-used data augmentations in DETR frameworks~\cite{carion2020end,zhu2020deformable,zhang2022dino,yang2023explicit}, including random flip, random crop, and random resize with the short sides in $[480, 800]$ and the long side less or equal to $1333$. We use the AdamW optimizer\cite{kingma2014adam,loshchilov2017decoupled} with the weight decay $1 \times 10^{-4}$ and train $60$ epochs and $80$ epochs on MS COCO~\cite{lin2014microsoft} and CrowdPose~\cite{li2019crowdpose}, respectively. We adopt a total batch size of $16$, and set the base learning rate as $1 \times 10^{-4}$. The base learning rate for the backbone is $1 \times 10^{-5}$ following the DETR frameworks. The learning rates are decayed at the $50$-th epoch and $70$-th by a factor of $0.1$ for MS COCO and CrowdPose, respectively. During testing, we resize the input images with their short sides being $800$ and long sides less or equal to $1333$.

\subsection{Main Results}
Our purpose is to build a simple baseline for end-to-end multi-person pose estimation. Thus, we mainly compare our Group Pose with previous end-to-end frameworks, including PETR~\cite{shi2022end}, QueryPose~\cite{xiao2022querypose}, and ED-Pose~\cite{yang2023explicit}. Besides, to show the effectiveness of our method, we also add comparisons with non-end-to-end frameworks, such as top-down~\cite{he2017mask,li2021pose}, bottom-up~\cite{cheng2020higherhrnet,geng2021bottom,luo2021rethinking,xue2022learning}, and one-stage methods~\cite{tian2019directpose,zhou2019objects,mao2021fcpose,shi2021inspose}.
%-------------------------------------------------------------------------
% main tables on COCO test
\begin{table*}[t]
\centering
\setlength{\tabcolsep}{9.2pt}
\renewcommand{\arraystretch}{1.25}
\footnotesize
\captionsetup{font=small}
\caption{\textbf{Comparisons with state-of-the-art methods on MS COCO \texttt{test-dev2017} dataset.} Notations are consistent with Table~\ref{tab:table1}.}
\begin{tabular}{c|c|l|c|l|l|ccc|cc}
\multicolumn{3}{c|}{Method} & Ref & Backbone & Loss & AP & AP$_{50}$ & AP$_{75}$ & AP$_{M}$ & AP$_{L}$ \\
\shline
\multirow{12}{*}{\rotatebox{90}{Non-End-to-End}} & \multirow{4}{*}{\rotatebox{90}{Top-Down}} & Mask R-CNN \cite{he2017mask} & CVPR 17 & ResNet-50 & HM  & $ 63.9 $ & $ 87.7 $ & $ 69.9 $ & $ 59.7 $ & $ 71.5 $ \\
& & Mask R-CNN \cite{he2017mask} & CVPR 17 & ResNet-101 & HM  & $ 64.3 $ & $ 88.2 $ & $ 70.6 $ & $ 60.1 $ & $ 71.9 $ \\
& & PRTR$^\dagger$ \cite{li2021pose} & CVPR 21 & ResNet-101 & KR  & $ 68.8 $ & $ 89.9 $ & $ 76.9 $ & $ 64.7 $ & $ 75.8 $ \\
& & PRTR$^\dagger$ \cite{li2021pose} & CVPR 21 & HRNet-w32 & KR  & $ 71.7 $ & $ 90.6 $ & $ 79.6 $ & $ 67.6 $ & $ 78.4 $ \\
\cline{2-11}
& \multirow{4}{*}{\rotatebox{90}{Bottom-Up}} & HrHRNet$^\dagger$ \cite{cheng2020higherhrnet} & CVPR 20 & HRNet-w32 & HM & $ 66.4 $ & $ 87.5 $ & $ 72.8 $ & $ 61.2 $ & $ 74.2 $ \\
& & DEKR$^\dagger$ \cite{geng2021bottom}      & CVPR 21 & HRNet-w32 & HM & $ 67.3 $ & $ 87.9 $ & $ 74.1 $ & $ 61.5 $ & $ 76.1 $ \\
& & SWAHR$^\dagger$ \cite{luo2021rethinking}  & CVPR 21 & HRNet-w32 & HM & $ 67.9 $ & $ 88.9 $ & $ 74.5 $ & $ 62.4 $ & $ 75.5 $ \\
& & LOGO-CAP$^\dagger$ \cite{xue2022learning} & CVPR 22 & HRNet-w32 & HM & $ 68.2 $ & $ 88.7 $ & $ 74.9 $ & $ 62.8 $ & $ 76.0 $ \\
\cline{2-11}
& \multirow{4}{*}{\rotatebox{90}{One-Stage}} & DirectPose \cite{tian2019directpose} & $-$ & ResNet-50 & KR & $ 62.2 $ & $ 86.4 $ & $ 68.2 $ & $ 56.7 $ & $ 69.8 $ \\
& & CenterNet$^\dagger$ \cite{zhou2019objects} & $-$ & Hourglass-104 & KR+HM & $ 63.0 $ & $ 86.8 $ & $ 69.6 $ & $ 58.9 $ & $ 70.4 $ \\
& & FCPose \cite{mao2021fcpose} & CVPR 21 & ResNet-50 & KR+HM & $ 64.3 $ & $ 87.3 $ & $ 71.0 $ & $ 61.6 $ & $ 70.5 $ \\
& & InsPose \cite{shi2021inspose} & ACM MM 21 & ResNet-50 & KR+HM & $ 65.4 $ & $ 88.9 $ & $ 71.7 $ & $ 60.2 $ & $ 72.7 $ \\
\shline
\multirow{9}{*}{\rotatebox{90}{End-to-End}} & \multirow{5}{*}{\rotatebox{90}{Previous Works}} & PETR \cite{shi2022end} & CVPR 22  & ResNet-50 & HM+KR  & $ 67.6 $ & $ 89.8 $ & $ 75.3 $ & $ 61.6 $ & $ 76.0 $ \\ 
& & PETR \cite{shi2022end} & CVPR 22 & Swin-L    & HM+KR & $ 70.5 $ & $ 91.5 $ & $ 78.7 $ & $ 65.2 $ & $ 78.0 $ \\
& & QueryPose \cite{xiao2022querypose} & NeurIPS 22 & Swin-L    & BR+RLE & $ 72.2 $ & $ 92.0 $ & $ 78.8 $ & $ 67.3 $ & $ 79.4 $ \\
& & ED-Pose \cite{yang2023explicit} & ICLR 23 & ResNet-50 & BR+KR & $ 69.8 $ & $ 90.2 $ & $ 77.2 $ & $ 64.3 $ & $ 77.4 $ \\
& & ED-Pose \cite{yang2023explicit} & ICLR 23 & Swin-L    & BR+KR & $ 72.7 $ & $ 92.3 $ & $ 80.9 $ & $ 67.6 $ & $ 80.0 $ \\
\cline{2-11}
& \multirow{3}{*}{\rotatebox{90}{Ours}} & GroupPose   & $-$     & ResNet-50 & KR     & $ 70.2 $ & $ 90.5 $ & $ 77.8 $ & $ 64.7 $ & $ 78.0 $ \\
& & GroupPose   & $-$     & Swin-T    & KR     & $ 72.1 $ & $ 91.4 $ & $ 79.9 $ & $ 66.7 $ & $ 79.5 $ \\
& & GroupPose   & $-$     & Swin-L    & KR     & $\bf 72.8 $ & $\bf 92.5 $ & $\bf 81.0 $ & $\bf 67.7 $ & $\bf 80.3 $ \\
\end{tabular}
\label{tab:table2}
\vspace{-3mm}
\end{table*}
%-------------------------------------------------------------------------

\vspace{1mm}
\noindent \textbf{Comparisons with end-to-end frameworks on COCO.} Table~\ref{tab:table1} and Table~\ref{tab:table2} present the comparisons on COCO \texttt{val2017} set and \texttt{test-dev} set. Results show that Group Pose outperforms PETR~\cite{shi2022end}, QueryPose~\cite{xiao2022querypose}, and ED-Pose~\cite{yang2023explicit} consistently. 

On COCO \texttt{val2017}, Group Pose surpasses PETR and QueryPose by over a significant $3.0$ AP with ResNet-50~\cite{he2016deep} and the gaps remain $1.5+$ AP with a strong backbone, Swin-Large~\cite{liu2021swin}. When comparing with the recently proposed ED-Pose~\cite{yang2023explicit}, which transfers pose estimation to a keypoint box detection problem and combines human box detection task, Group Pose can also exceed it with non-negligible margins. Moreover, unlike these methods that use complex decoders and add extra supervisions, {\em e.g.}, extra heatmap or box supervisions, our Group Pose only use a simple decoder and are trained only with keypoint regression targets. The above evidences indicate that complex design for end-to-end multi-person pose estimation may not be necessary. 

On COCO \texttt{test-dev}, our Group Pose achieves $70.2$ AP, $72.1$ AP, and $72.8$ AP with ResNet-50~\cite{he2016deep}, Swin-Tiny~\cite{liu2021swin}, and Swin-Large~\cite{liu2021swin} as the backbone. Compared with other end-to-end frameworks, similar trends are observed with the ones on COCO \texttt{val2017}.

\vspace{1mm}
\noindent \textbf{Comparisons with end-to-end frameworks on CrowdPose.} To further demonstrate the effectiveness of Group Pose, we provide comparisons with previous end-to-end frameworks on the challenging CrowdPose dataset~\cite{li2019crowdpose}. Table~\ref{tab:table3} reports the results of PETR~\cite{shi2022end}, QueryPose~\cite{xiao2022querypose}, ED-Pose~\cite{yang2023explicit}, and our Group Pose with the same backbone Swin-Large~\cite{liu2021swin}. Overall, Group Pose achieve $74.1$ AP, performing the best over all methods. 

Moreover, it is interesting when we compare the AP scores with easy, medium, and hard crowding levels of different models. PETR~\cite{shi2022end} performs worse with the easy crowding level while giving the second best results with the hard level. ED-Pose~\cite{yang2023explicit} performs worst with the hard crowding level. We conjecture that this phenomenon is caused by the differences in their decoders, {\em e.g.}, human instances crowded with each other have similar boxes, which challenges the human keypoint decoder. Our Group Pose, which uses $K$ keypoint queries and one instance query for each human instance and considers different interactions among queries, can achieve reasonable good results with easy, medium, and hard crowding levels.

\vspace{1mm}
\noindent \textbf{Comparisons with various non-end-to-end frameworks on COCO.} We also compare Group Pose with representative non-end-to-end frameworks in Table~\ref{tab:table1} and Table~\ref{tab:table2}. Group Pose can easily outperform previous bottom-up methods~\cite{cheng2020higherhrnet,geng2021bottom,luo2021rethinking,xue2022learning} and one-stage methods~\cite{tian2019directpose,zhou2019objects,mao2021fcpose,shi2021inspose}. For example, Group Pose surpasses the recent proposed LOGO-CAP~\cite{xue2022learning} with flipping test by over $2.0$ AP on both COCO \texttt{val2017} and \texttt{test-dev}, even with a smaller backbone (ResNet-50~\cite{he2016deep}) than it (HRNet-w32~\cite{sun2019deep}). Group Pose is more concise and precise than those bottom-up and one-stage methods. Surprisingly, Group Pose can also beat previous top-down methods like PRTR~\cite{li2021pose} and Poseur~\cite{mao2022poseur}. Given the simple design and the end-to-end property, our Group Pose can serve as a simple baseline for pursuing higher performance on multi-person pose estimation.
%-------------------------------------------------------------------------
% main table on CrowedPose
\begin{table}[t]
\centering
\setlength{\tabcolsep}{7pt}
\renewcommand{\arraystretch}{1.4}
\captionsetup{font=small}
\caption{\textbf{Comparisons with state-of-the-art methods on CrowdPose \texttt{test} dataset.} Swin-L is adopted as the backbone. Other notations are consistent with Tabel~\ref{tab:table1}.}
\resizebox{\linewidth}{!}{
\begin{tabular}{l|l|ccc|ccc}
Method & Loss & AP & AP$_{50}$ & AP$_{75}$ & AP$_{E}$ & AP$_{M}$ & AP$_{H}$ \\
\shline
PETR~\cite{shi2022end}        & HM+KR & $ 71.6 $ & $ 90.4 $ & $ 78.3 $ & $ 77.3 $ & $ 72.0 $ & $ 65.8 $ \\ 
QueryPose~\cite{xiao2022querypose}   & BR+RLE & $ 72.7 $ & $\bf 91.7 $ & $ 78.1 $ & $ 79.5 $ & $ 73.4 $ & $ 65.4 $ \\ 
ED-Pose~\cite{yang2023explicit}      & BR+KR & $ 73.1 $ & $ 90.5 $ & $ 79.8 $ & $ 80.5 $ & $ 73.8 $ & $ 63.8 $ \\ 
\hline
GroupPose   & KR     & $\bf 74.1 $ & $ 91.3 $ & $\bf 80.4 $ & $\bf 80.8 $ & $\bf 74.7 $ & $\bf 66.4 $ \\ 
\end{tabular}}
\label{tab:table3}
\vspace{-3mm}
\end{table}
%-------------------------------------------------------------------------
%-------------------------------------------------------------------------
\begin{table*}[t]
\centering
\captionsetup{font=small}
\caption{\textbf{Ablation experiments for Group Pose.} Evaluated on MS COCO \texttt{val2017}. Default settings are marked in \colorbox{lightgray!20}{gray}.}
\vspace{-2mm}
    \begin{subtable}[t]{0.3\textwidth}
        \centering
        \renewcommand\arraystretch{1.0}
        \caption{\textbf{Query designs for human instances.} Both instance (inst) and keypoint (kpt) queries are essential in Group Pose, especially the keypoint ones.}
        \resizebox{0.85\linewidth}{!}{
        \begin{tabular}{c|ccc} 
        query types & AP & AP$_{M}$ & AP$_{L}$ \\
        \shline
        \cellcolor{lightgray!20}{inst $\&$ kpt} &\cellcolor{lightgray!20}{$\bf 72.0 $} &\cellcolor{lightgray!20}{$\bf 66.8 $} &\cellcolor{lightgray!20}{$\bf 79.7 $} \\
        only kpt & $ 71.2 $ & $ 66.0 $ & $ 79.1 $ \\
        only inst & $ 64.5 $ & $ 61.1 $ & $ 69.9 $ \\
        \end{tabular}}
        \label{tab:ablationquerydesign}
     \end{subtable}
    \hfill
    \begin{subtable}[t]{0.35\textwidth}
        \centering
        \renewcommand\arraystretch{1.0}
        \caption{\textbf{Benefits of the instance query.} On the model only with kpt queries, the inst query is first added but not for classification (cls). Then we use the inst query for the cls task.}
        \resizebox{0.95\linewidth}{!}{
        \begin{tabular}{cc|ccc}
        query types & cls task & AP & AP$_{M}$ & AP$_{L}$ \\
        \shline
        only kpt & avg kpt & $71.2 $ & $ 66.0 $ & $ 79.1 $ \\
        inst $\&$ kpt & avg kpt & $ 71.7 $ & $ 66.8 $ & $ 79.3 $ \\
        \cellcolor{lightgray!20}{inst $\&$ kpt} & \cellcolor{lightgray!20}{inst} & \cellcolor{lightgray!20}{$\bf 72.0 $} & \cellcolor{lightgray!20}{$\bf 66.8 $} &\cellcolor{lightgray!20}{$\bf 79.7 $} \\ 
        \end{tabular}
        }
        \label{tab:ablationinstancequery}
     \end{subtable}
      \hfill
     \begin{subtable}[t]{0.3\textwidth}
        \centering
        \renewcommand\arraystretch{1.0}
        \caption{\textbf{Number of instances.} When the number is smaller than $100$, increasing instance numbers provide gains. $100$ is set by default for the number of instances.}
        \resizebox{0.80\linewidth}{!}{
        \begin{tabular}{c|ccc}
        \#instance & AP & AP$_{M}$ & AP$_{L}$ \\
        \shline
        $50$  & $ 71.4 $ & $ 66.4 $ & $ 79.1 $ \\
        \cellcolor{lightgray!20}{$100$} &\cellcolor{lightgray!20}{$\bf 72.0 $} &\cellcolor{lightgray!20}{$ 66.8 $} &\cellcolor{lightgray!20}{$\bf 79.7 $} \\
        $200$ & $\bf 72.0 $ & $\bf 66.9 $ & $\bf 79.7 $ \\
        \end{tabular}}
        \label{tab:ablationnuminstance}
     \end{subtable}
     \hfill
     \begin{subtable}[t]{0.48\textwidth}
        \centering
        \renewcommand\arraystretch{1.0}
        \vspace{2mm}
        \caption{\textbf{Self-attention implementations.} In the self-attention module, removing the across-instance interactions for queries with different types with attention (attn) mask or group self-attentions eases the optimization and improves the performance.}
        \resizebox{0.95\linewidth}{!}{
        \begin{tabular}{cc|ccc}
        self-attention implementations & w/ attn mask & AP & AP$_{M}$ & AP$_{L}$ \\
        \shline
        standard self-attention & $\times$ & $ 69.4 $ & $ 64.0 $ & $ 77.5 $ \\
        standard self-attention & \checkmark & $ 70.7 $ & $ 65.5 $ & $ 78.4 $ \\
        \cellcolor{lightgray!20}{group self-attentions} & \cellcolor{lightgray!20}{$\times$} &\cellcolor{lightgray!20}{$\bf 72.0 $} &\cellcolor{lightgray!20}{$\bf 66.8 $} &\cellcolor{lightgray!20}{$\bf 79.7 $} \\
        \end{tabular}}
        \label{tab:ablationsaimplementations}
     \end{subtable}
     \hfill
    \begin{subtable}[t]{0.48\textwidth}
        \centering
        \renewcommand\arraystretch{1.0}
        \vspace{2mm}
        \caption{\textbf{Group self-attentions.} Both types of self-attentions are important in group self-attentions. We also conduct an experiment by removing both within-instance and across-instance self-attentions. The performance further drops to $65.3$ AP.}
        \resizebox{0.95\linewidth}{!}{
        \begin{tabular}{c|ccc}
        group self-attentions & AP & AP$_{M}$ & AP$_{L}$ \\
        \shline
        \cellcolor{lightgray!20}{wthin-instance $\&$ across-instance self-attentions} &\cellcolor{lightgray!20}{$\bf 72.0 $} &\cellcolor{lightgray!20}{$\bf 66.8 $} &\cellcolor{lightgray!20}{$\bf 79.7 $} \\
        within-instance self-attentions & $ 66.3 $ & $ 63.8 $ & $ 70.8 $ \\
        across-instance self-attentions & $ 67.4 $ & $ 63.0 $ & $ 74.4 $ \\
        \end{tabular}}
        \label{tab:ablationgroupsa}
     \end{subtable}
\vspace{-3mm}
\end{table*}
%-------------------------------------------------------------------------

\subsection{Ablation Study}
\label{sec:ablation}
We run a number of ablation experiments to verify the effectiveness of key elements in our Group Pose. We adopt ResNet-50~\cite{he2016deep} as the backbone. Unless specified, we report the results on COCO \texttt{val2017} with $60$ epochs training.

\vspace{1mm}
\noindent \textbf{Ablation: query designs for human instance.} Table~\ref{tab:ablationquerydesign} gives the results of representing a human instance with different query designs. Both instance query and keypoint queries are essential in Group Pose. As the multi-person pose estimation is to predict human poses given an image, it is reasonable that keypoint queries themself can achieve good results, while removing keypoint queries gives a significant performance drop (from $72.0$ AP to $64.5$ AP). 

For the instance query, its benefits come from two aspects: (i) gather information within human instance and help model training and (ii) decouple the classification task and keypoint positions regression task. Table~\ref{tab:ablationinstancequery} shows the improvements brought by these two benefits.

\begin{figure}[t]
\centering
\begin{tikzpicture}[font=\footnotesize]
\begin{axis}[
legend columns=1, 
legend style={at={(0.77,0.56)},anchor=north, font=\tiny, draw=gray!20}, legend cell align={left},
y label style={at={(0.07,0.5)}},
ylabel={Average Precision (AP)}, ymajorgrids=true, xtick={0,12,24,36,48,60}, xticklabels={ $0$, $12$, $24$, $36$, $48$, $60$},
ytick={20, 30, 40, 50, 60, 70},
tick style={draw=none},
y label style={font=\footnotesize},
height=6cm,
width=9cm,
y post scale=0.6,
x post scale=0.95,
axis lines = left,
every outer y axis line/.style={draw=gray!40},
every outer x axis line/.style={draw=gray!40},
grid style={line width=.1pt, draw=gray!20},
xmin=1, xmax=62.5, ymax=75, ymin=30]

\addplot[line width=0.9pt, mark size=1.5pt, draw=selfyellow, draw opacity=0.8]
table
{
X Y
1 6.051301071825758
2 22.741707725254805
3 33.16706940983396
4 38.04132382755256
5 43.9075125517343
6 47.34443735905278
7 49.667990282768784
8 50.52776219597791
9 53.37813608071287
10 53.883696310348014
11 56.20732140335877
12 58.46422631552509
13 58.483310975761846
14 59.04460008501473
15 60.38085748405434
16 61.26171461906233
17 61.061735532643766
18 61.6142066870212
19 62.91501509525625
20 62.99809257782076
21 63.47140230937476
22 63.476829690629
23 64.23397357216717
24 64.86501190936858
25 65.13904941448321
26 65.09596641530212
27 64.41432434712114
28 65.89479856983009
29 66.30987627588536
30 66.65674346824895
31 66.00487529113515
32 66.86908841007731
33 66.48276056382676
34 66.96157645309798
35 67.51766842580471
36 67.33724966109202
37 67.1549484171891
38 67.69015371821364
39 67.2973607533498
40 67.61900646725711
41 67.82699703339026
42 68.46065349635649
43 68.45670869957765
44 68.06512036438419
45 68.51216560943622
46 68.30344767624638
47 68.91252958775958
48 68.69006387556598
49 68.87120278641875
50 68.75667847317001
51 71.06207032781309
52 71.42906228183618
53 71.60353309325357
54 71.7098075240985
55 71.572155047628
56 71.77607520039489
57 71.75857525045038
58 71.87740584018466
59 71.9212413357623
60 71.95815487659951
};
\addlegendentry{group self-attentions}

\addplot[line width=0.9pt, mark size=1.5pt, draw=selfgreen, draw opacity=0.8]
table
{
X Y
1 3.5037216986247866
2 13.720727752734035
3 21.854766835741792
4 32.39481870209461
5 36.14706200598952
6 40.95899728294308
7 41.711414679041184
8 45.58721569637266
9 46.125385553137825
10 49.50849089660103
11 48.82187521437339
12 52.274269471621125
13 52.138341817295164
14 52.778436421374096
15 55.172700559447726
16 55.17996524269266
17 55.19370804207884
18 56.40842541747391
19 58.64895336551428
20 58.741948309268544
21 59.00929598261991
22 59.28149750919456
23 59.55906193578841
24 60.67109905206768
25 60.72390834062783
26 60.506682195325425
27 61.131081132463216
28 61.563295225750636
29 61.47151168351118
30 62.55303140347167
31 61.887051144231386
32 63.47166831036284
33 64.06865702819599
34 63.31182859797754
35 63.54304898543878
36 64.42579873890077
37 64.78443118204919
38 64.81626749917706
39 65.02575616898602
40 64.2697078018873
41 64.9289427268899
42 65.42697912694803
43 64.50839910124776
44 66.48020752458945
45 65.8561671364225
46 65.9348008043468
47 66.05799592477119
48 66.24326518125075
49 66.21072827270076
50 66.34371458912342
51 68.69729123583332
52 69.03275417320344
53 69.23127000424869
54 69.31966832453465
55 69.18524462915803
56 69.3530263741813
57 69.31917522627542
58 69.39577704662373
59 69.63493791248024
60 69.429481577982
};
\addlegendentry{standard self-attention}

\addplot[line width=0.9pt, mark size=1.5pt, draw=selfblue, draw opacity=0.8]
table
{
X Y
1 7.370295983028219
2 15.565337668040533
3 25.566792005163606
4 33.005565428405355
5 39.626915383362224
6 41.92006847323195
7 45.41140170981193
8 49.09264782648622
9 51.006788932445104
10 50.75993905136149
11 53.03369072000124
12 53.576981361147894
13 55.58410834951314
14 56.093236452201566
15 58.15721447108434
16 58.10110675014023
17 58.74900406463237
18 59.78588953491215
19 59.64812984168216
20 61.58970601066544
21 61.03472202110017
22 61.4526976832278
23 62.332960210607666
24 62.44456150352731
25 63.151206095828435
26 63.74430302739208
27 63.81748733400047
28 62.5308462211781
29 63.612310346353965
30 62.991877278811415
31 64.6276872879141
32 65.77319299584181
33 64.75790673275917
34 64.79965071152478
35 64.87981626808192
36 64.78892937106077
37 66.13572311406091
38 66.58390548520669
39 66.35791682573151
40 65.67242085239438
41 66.43929244086243
42 65.41826460789673
43 66.07106129620497
44 67.1989588796009
45 67.33440945090774
46 67.0741755958069
47 66.34756441764036
48 66.92241506702847
49 67.08016857803447
50 67.55165273236776
51 70.00028491091945
52 70.18213999850083
53 70.34979721202542
54 70.47417462050075
55 70.34421143723377
56 70.56007275752046
57 70.58871466602075
58 70.51695771401256
59 70.6987150122487
60 70.4748546165723
};
\addlegendentry{standard self-attention w/ mask}

\end{axis}
\node [above right] at (6.65,2.4) {\tiny 72.0};
\node [above right] at (6.65,2.2) {\tiny 70.7};
\node [above right] at (6.65,2.0) {\tiny 69.4};

\end{tikzpicture}
\captionsetup{font=small}
\caption{\textbf{Faster and better convergence.} 
The $x$-axis corresponds to \#epoch,
and the $y$-axis corresponds to AP score. One can see that group self-attentions converge faster and better than other self-attention implementations.
}
\label{fig:comparisoncurves}
\vspace{-3mm}
\end{figure}
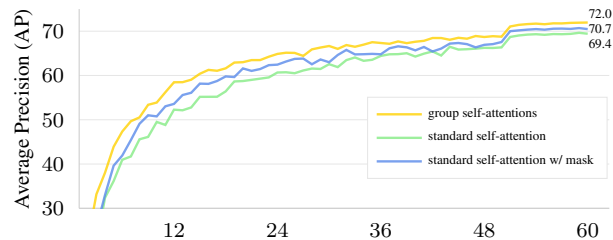

\vspace{1mm}
\noindent \textbf{Ablation: self-attention implementations.} Table~\ref{tab:ablationsaimplementations} provides the comparisons between different self-attentions in the decoder layers, including (i) standard self-attention over all the $N \times (K + 1)$ queries (Figure~\ref{fig:interaction} (a)), (ii) standard self-attention with an attention mask (Figure~\ref{fig:interaction} (b)), which masks out the across-instance interactions for
queries with different types, and (iii) our group self-attentions (Figure~\ref{fig:interaction} (c)). Results validate that the across-instance interactions for
queries with different types are not directly helpful for the multi-person pose estimation task. Removing this type of interactions eases the model optimization, helps model converge faster, thus improves the performances. The comparisons on convergence curves are given in Figure~\ref{fig:comparisoncurves}. Besides, we find that it is better to separately perform different types of interactions, the within-instance and same-type across-instance interactions, with different parameters. Table~\ref{tab:ablationsaimplementations} shows that our sequential implementation group self-attentions bring a $1.3$ AP gain over the parallel implementation of standard self-attention with an attention mask.
%-------------------------------------------------------------------------
%-------------------------------------------------------------------------
%-------------------------------------------------------------------------
\begin{figure*}[t]
\centering
\includegraphics[width=0.98\textwidth]{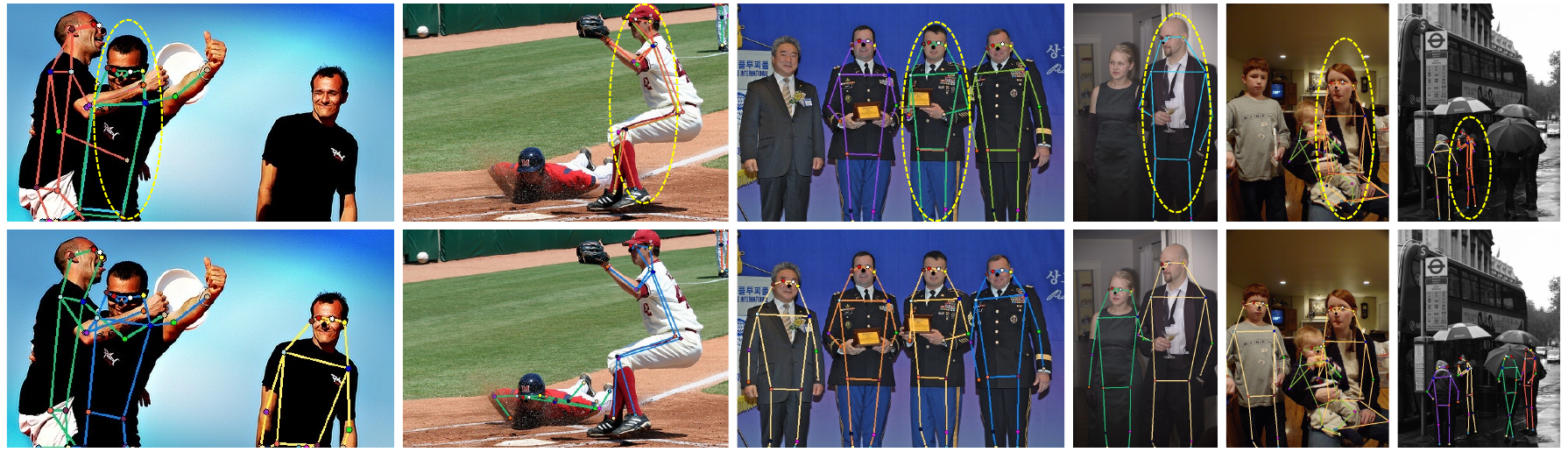}
\captionsetup{font=small}
\caption{\textbf{Comparison of removing duplicated predictions.} We visualize the predicted poses of Group Pose (second row), and Group Pose without same-type across-instance self-attentions (first row) according to the number of objects. Group Pose produces less duplicated pose predictions. The yellow dashed ellipse indicates the duplicated predicted human instance, leading to a mismatch of other instances. Best view in zoom in.}
\label{fig:visduplicatepredictions}
\vspace{-3mm}
\end{figure*}
%-------------------------------------------------------------------------

\vspace{1mm}
\noindent \textbf{Ablation: group self-attentions.} We ablate the effects of $N$ within-instance self-attentions over $K$ keypoint queries and one instance query and $(K + 1)$ across-instance self-attentions over $N$ queries of the same type in Table~\ref{tab:ablationgroupsa}. Large performance gaps (over $-5.0$ AP) are observed when we remove either of these two group self-attentions. Both the interactions are important for multi-person pose estimation. The within-instance self-attentions explore kinematic relations and gather information for scoring pose predictions, which help the model predict precise keypoint positions for human poses. The same-type across-instance self-attentions collect information from the same-type of keypoint queries or instance queries, removing duplicate predictions for poses among human instances. As shown in Figure~\ref{fig:visduplicatepredictions}, model without the same-type across-instance self-attentions produces more duplicated pose predictions for the same human instance than Group Pose.

\vspace{1mm}
\noindent \textbf{Ablation: number of instances.} Table~\ref{tab:ablationnuminstance} gives how pre-defined number of human instances affect the results. We find $100$ human instances are enough and can achieve comparable results with $200$ human instances. We set the number of human instances as $100$ by default. Thus, Group Pose has $100\times (17 + 1)$ queries on COCO~\cite{lin2014microsoft} and contains $100 \times (14 + 1)$ queries on CrowdPose~\cite{li2019crowdpose}.

\subsection{More Analysis}
We provide more analysis about our Group Pose in this section, including the analysis on model convergence and the analysis on model inference speed, detailed next.
%-------------------------------------------------------------------------
\begin{table}[t]
\centering
\setlength{\tabcolsep}{5pt}
\renewcommand{\arraystretch}{1.5}
\footnotesize
\captionsetup{font=small}
\caption{\textbf{Analysis on model convergence.} Group Pose without human detection already can outperform ED-Pose~\cite{yang2023explicit}. Besides, Group Pose can also benefit from better human instance initialization with human detection. We report AP on COCO \texttt{val2017} dataset. `Det Dec' = human detection decoder. The results of ED-Pose are from the original paper \cite{yang2023explicit}.}
\begin{tabular}{l|c|ccccc}
Method & w/ Det Dec & 12e & 24e & 36e & 48e & 60e \\
\shline
ED-Pose~\cite{yang2023explicit}  & $ \checkmark $ & $ 60.5 $ & $ 67.5 $ & $ 69.7 $ & $ 70.8 $ & $ 71.6 $ \\ 
\hline
GroupPose  & $\times$  & $\bf 61.0 $ & $\bf 67.6 $ & $\bf 70.1 $ & $\bf 71.4 $ & $\bf 72.0 $ \\ 
GroupPose  & \checkmark  & $\bf 61.4 $ & $\bf 68.1 $ & $\bf 70.3 $ & $\bf 71.6 $ & $\bf 72.2$ \\ 
\end{tabular}
\label{tab:convergencecomparisons}
\vspace{-4mm}
\end{table}
%-------------------------------------------------------------------------

\vspace{1mm}
\noindent \textbf{Analysis on model convergence.} A common wisdom about training models is to provide a good initialization and then refine the model based on it. ED-Pose~\cite{yang2023explicit} splits the multi-person pose estimation task into two sub-processes, which first detect human instances with a human detection decoder and then use a human-to-keypoint detection decoder for predicting human poses. The human detection decoder gives good initialization for human instances, which can help ED-Pose learn and converge faster. Although our Group Pose adopts a different design for transformer decoder, it can also be built upon a human detection decoder and enjoys the benefits brought by better initialization of human instances. Based on this observation, we build a variant of our Group Pose, whose transformer decoder consists of $2$ human detection decoder layers and $4$ simple transformer decoder layers for multi-person pose estimation, following ED-Pose~\cite{yang2023explicit}.
%-------------------------------------------------------------------------
\begin{table}[ht]
\centering
\setlength{\tabcolsep}{10pt}
\renewcommand{\arraystretch}{1.15}
\footnotesize
\captionsetup{font=small}
\caption{\textbf{Analysis on model inference speed.} Frames per second (FPS) and inference time (Time) are measured with ResNet-50 and different image resolutions on one NVIDIA A100 GPU.}
\begin{tabular}{l|c|c|c}
Method & Input Resolution & FPS $\uparrow$ & Time [ms] $\downarrow$ \\
\shline
\multirow{2}{*}{PETR~\cite{shi2022end}} & $ 480 \times 800 $  & $20.0$ & $50$ \\ 
                                                      & $ 800 \times 1333 $ & $12.1$ & $83$ \\ 
\hline
\multirow{2}{*}{QueryPose~\cite{xiao2022querypose}} & $ 480 \times 800 $  & $19.0$ & $56$ \\ 
                                                       & $ 800 \times 1333 $ & $13.4$ & $75$ \\ 
\hline
\multirow{2}{*}{ED-Pose~\cite{yang2023explicit}}  & $ 480 \times 800 $  & $42.4$ & $24$ \\ 
                                                      & $ 800 \times 1333 $ & $24.7$ & $40$ \\ 
\shline 
\multirow{2}{*}{GroupPose} & $ 480 \times 800 $  & $\bf 68.6$ & $\bf 15$ \\ 
                                                      & $ 800 \times 1333 $ & $\bf 31.3$ & $\bf 32$ \\ 
\end{tabular}
\label{tab:inferencespeed}
\vspace{-4mm}
\end{table}
%-------------------------------------------------------------------------

Table~\ref{tab:convergencecomparisons} shows the AP scores of three models, ED-Pose, Group Pose, and Group Pose with human detection decoder, under different training schedules. Group Pose already shows superior results than ED-Pose with $12$ epochs, $24$ epochs, $36$ epochs, $48$ epochs, and $60$ epochs training, even without the help of human detection decoder. Moreover, the comparison between Group Pose and Group Pose with human detection decoder verifies that good initialization for human instances leads to faster convergence and better results.

\vspace{1mm}
\noindent \textbf{Analysis on model inference speed.} Table~\ref{tab:inferencespeed} provides the comparisons of inference time and FPS among end-to-end frameworks, including PETR~\cite{shi2022end}, QueryPose~\cite{xiao2022querypose}, ED-Pose~\cite{yang2023explicit}, and our Group Pose. Simple designs usually show efficiency. Results show that our simple transformer decoder can be faster than complex decoders in previous end-to-end frameworks. Even with an image in a large size $800\times 1333$, our Group Pose can also achieve real-time speed (above $30$ FPS) on a single A$100$ GPU.

\section{Conclusion}
In this paper, we present a simple baseline, Group Pose, for end-to-end multi-person pose estimation. With simple designs for queries and decoder self-attentions, the approach outperforms previous end-to-end frameworks while being faster. Besides, the transformer decoder in Group Pose is also flexible, which can be built solely or upon human detection decoders. We hope Group Pose can provide insights for exploring concise and effective end-to-end multi-person pose estimation frameworks.

\noindent\textbf{Acknowledgements.} This work was supported in part by the National Key R\&D Program of China (No.2021ZD0112100), National NSF of China (No.U1936212, No.62120106009, No.62261160653).

{\small
\bibliographystyle{ieee_fullname}
\bibliography{main_arxiv}
}

%%%%%%%%%%%%%%%%%%%%%
\clearpage
\begin{appendices}
\section{Appendix}
\subsection{More details and analysis on inference speed}
We measure different end-to-end frameworks with ResNet-50 backbone on a single NVIDIA A100 GPU. To avoid the speed limitation of I/O and randomness of image augmentation, we pre-load test images with one fixed resolution ({\em e.g.}, $480 \times 800$ or $800 \times 1333$) into the GPU memory to remove the time cost of inputs pre-processing. Then, we directly send them into different inference models for a fair and precise comparison. Thus, the reported inference time (Time) in Table~\ref{tab:inferencespeed} is faster than their original papers of PETR~\cite{shi2022end}, QueryPose~\cite{xiao2022querypose} and ED-Pose~\cite{yang2023explicit}. 

Group Pose is faster than previous end-to-end frameworks with complex decoders. This can be explained by that Group Pose only contains a simple transformer decoder, thus eliminating some extra intermediate processes, {\em e.g.}, an additional query selection\footnote{Called `fine human query selection' in their paper.} in ED-Pose~\cite{yang2023explicit}.

\subsection{Ablation on across-instance interactions}
Group Pose captures $(K+1)$ across-instance interactions over $N$ queries of the same type, including one instance type and $K$ keypoint types, and the interaction designs are to be explored. With the same basic setting in Section \textcolor{red}{4.3}, the following table includes the relevant ablations:
\vspace{-2mm}
\begin{table}[h]
\centering
\setlength{\tabcolsep}{7pt}
\resizebox{0.9\linewidth}{!}{
\begin{tabular}{c|ccc}
across-instance interactions & AP & AP$_{M}$ & AP$_{L}$ \\
\shline
\hline
\cellcolor{lightgray!20}{inst-inst $\&$ kpt-kpt }  & \cellcolor{lightgray!20}{$\bf 72.0 $} & \cellcolor{lightgray!20}{$\bf 66.8$} & \cellcolor{lightgray!20}{$\bf 79.7 $} \\
only kpt-kpt    & $ 71.4 $ & $ 66.4 $ & $ 79.0 $ \\
only inst-inst    & $ 71.0 $ & $ 65.0 $ & $ 79.0 $ \\
\end{tabular}}
\vspace{-4mm}
\end{table}

\noindent Results validate that same-type across-instance interactions of both the instance (inst-inst) and keypoint (kpt-kpt) queries are essential in Group Pose. The across-instance interactions in Group Pose bring $+0.6$ and $+1.0$ AP gains over only modeling the keypoint queries and instance queries, suggesting the usefulness of promoting information aggregation of same-type queries, thus improving performance, as analyzed in Section \textcolor{red}{4.3}.

\subsection{Ablation on self-attentions implementations}
The proposed group self-attentions introduce two types of self-attentions, including within-instance and across-instance self-attentions. In practice, they are implemented with self-attention modules by calculating multiple attention maps in parallel. We ablate the effects of whether sharing the modules in the following table:

\vspace{-2mm}
\begin{table}[h]
\centering
\setlength{\tabcolsep}{7pt}
\resizebox{\linewidth}{!}{
\begin{tabular}{cc|ccc}
self-attention implementations & share & AP & AP$_{M}$ & AP$_{L}$ \\
\shline
\cellcolor{lightgray!20}{group self-attentions} & \cellcolor{lightgray!20}{$\times$} & \cellcolor{lightgray!20}{$\bf 72.0 $} & \cellcolor{lightgray!20}{$\bf 66.8$} & \cellcolor{lightgray!20}{$\bf 79.7 $} \\
group self-attentions &  \checkmark  & $ 71.5 $ & $ 66.3 $ & $ 79.4 $ \\ 
\end{tabular}}
\vspace{-4mm}
\end{table}

\noindent We can observe that sharing modules gives a $-0.5$ AP drop over the unshared one. This is mainly because the two types of self-attentions in Group Pose are responsible for gathering within-instance and across-instance information, respectively. Thus, it is reasonable that the unshared implementation can achieve a better result.

\subsection{Qualitative results on instance query}
\begin{figure*}[h]
\centering
\includegraphics[width=0.97\textwidth]{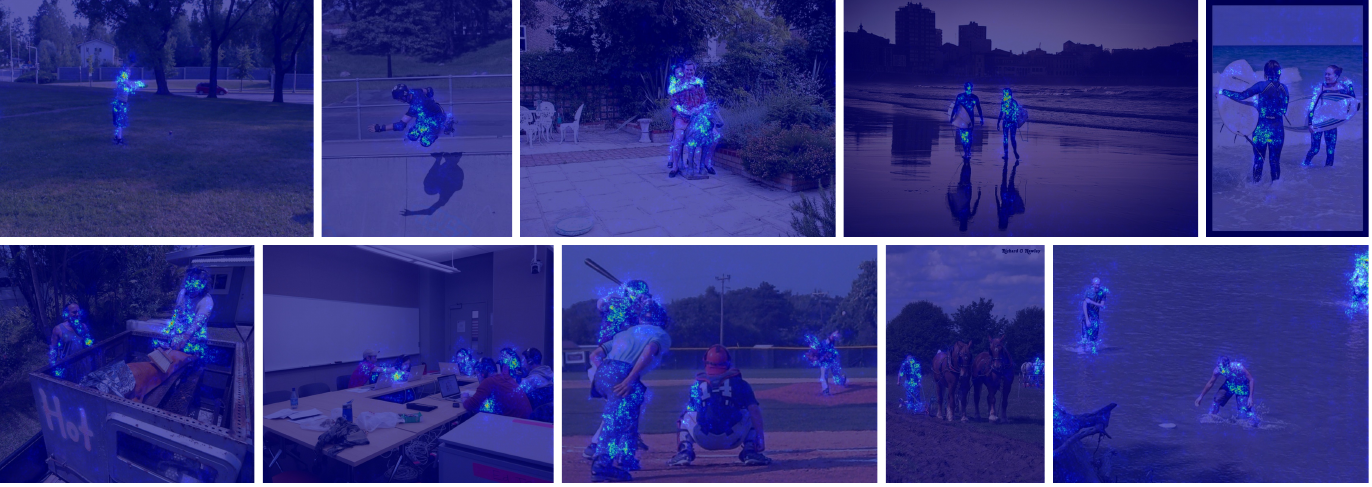}
\caption{\textbf{The gradient norm of instance query} with respect to each pixel in given images. The salient region is visualized by bright color. Best view in zoom in.}
\label{fig:visinstancequery}
\vspace{-3mm}
\end{figure*}

Group Pose directly utilizes instance query for classification to identify human instances. For studying what instance query looks at to give final results, we visualize the gradient norm of instance query with respect to each pixel in given images, as shown in Figure~\ref{fig:visinstancequery}. The gradient norm reflects the degree of change in final results due to each pixel interference, thus showing which pixels the instance query relies on for classification. The results show that the instance query in Group Pose looks at pixels inside human instances of given images even without human box supervision, thus accurately scoring the predicted poses.

\begin{figure*}[h]
\centering
\includegraphics[width=0.97\textwidth]{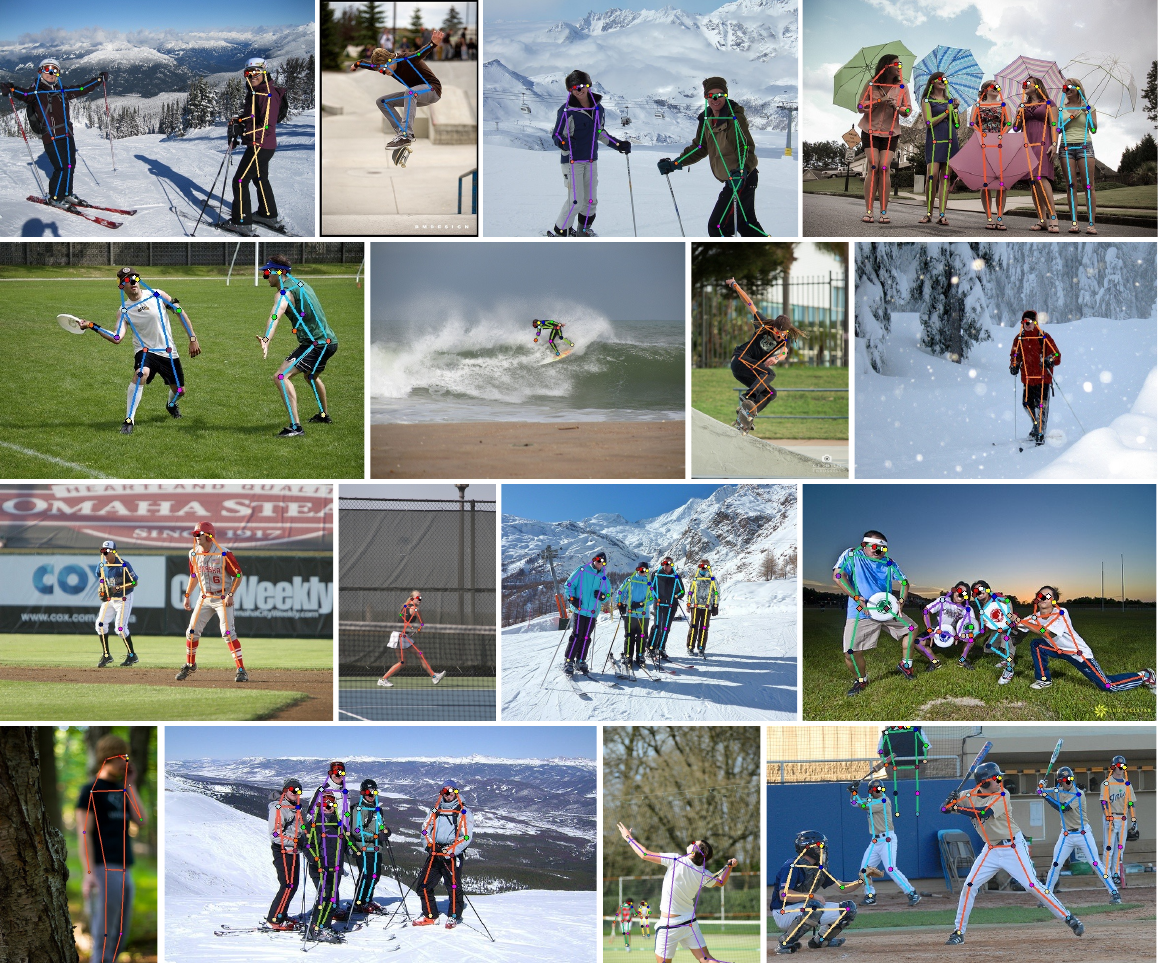}
\caption{\textbf{Visualization results of Group Pose on MS COCO.} Group Pose performs well with scale variations and pose deformations. Best view in color.}
\label{fig:viscoco}
\vspace{-6mm}
\end{figure*}

\subsection{Visualization results}
\begin{figure*}[h]
\centering
\includegraphics[width=0.97\textwidth]{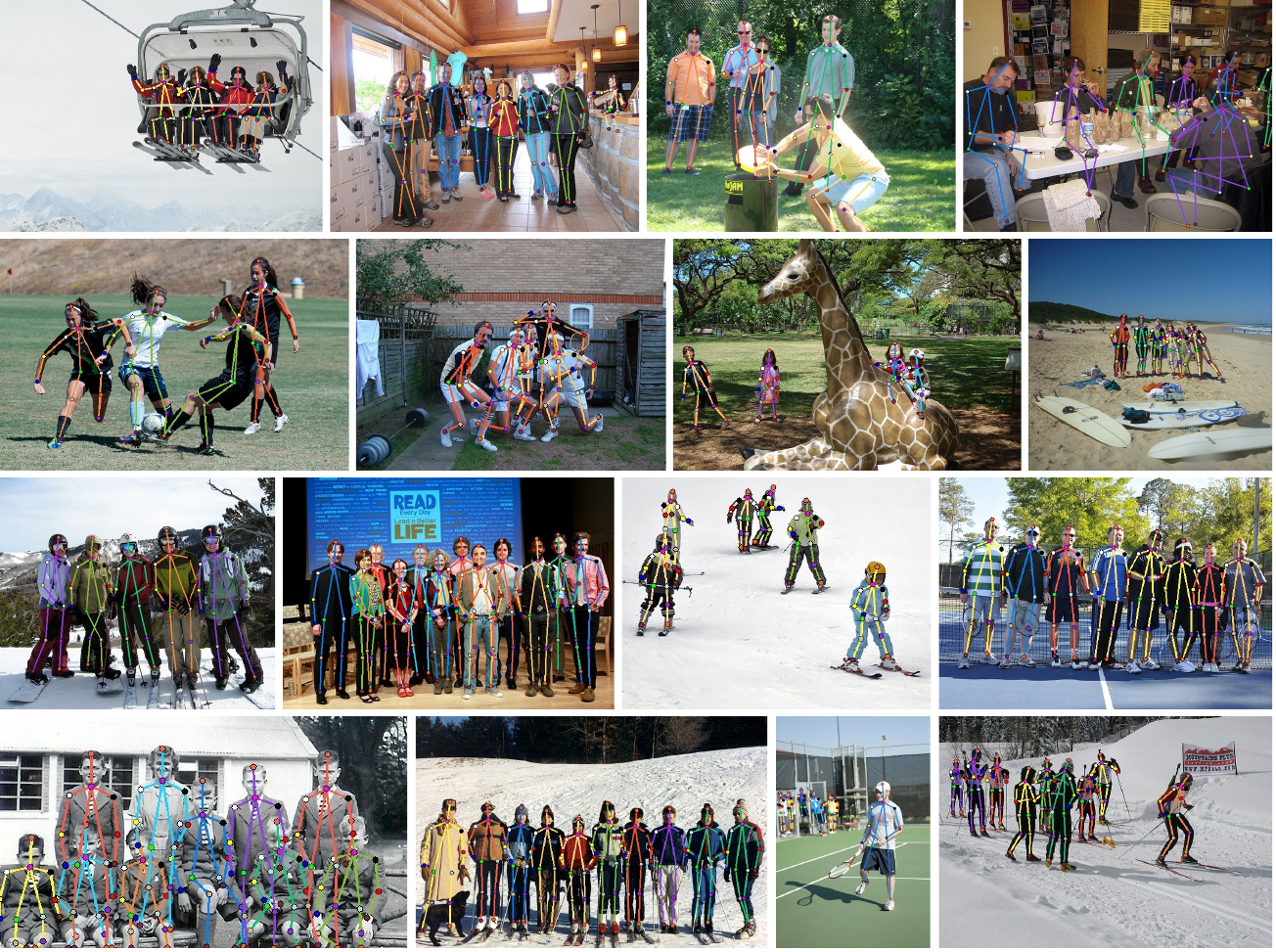}
\caption{\textbf{Visualization results of Group Pose on CrowdPose.} Group Pose is robust for challenging crowded and occluded scenes. Best view in color.}
\label{fig:viscrowdpose}
\end{figure*}

We visualize the predicted results on MS COCO in Figure~\ref{fig:viscoco} and CrowdPose in Figure~\ref{fig:viscrowdpose}. As can be observed, Group Pose performs well on a wide range of poses, including scale variations, motion blur, pose deformations, occlusion, and crowded scenes. The results demonstrate the effectiveness of our design of Group Pose for end-to-end multi-person pose estimation.

\subsection{Limitation} 
\begin{figure*}
\centering
\includegraphics[width=\linewidth]{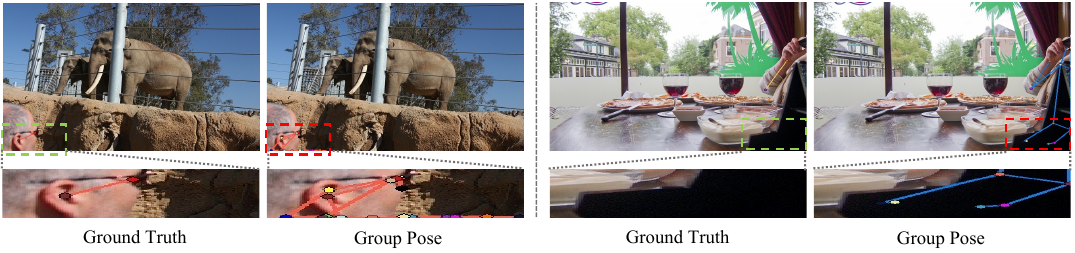}
\caption{\textbf{Visualization results on MS COCO images with small parts of human instance.} Model is based on ResNet-50. The red dashed box indicates difficulties in predicting unlabeled keypoints. Best view in color and zoom in.} 
\label{fig:limitaion}
\end{figure*}

Group Pose shows good results on benchmark datasets, while there are also some failure cases. We find that Group Pose has difficulties in the situation that only contains a small part ({\em e.g.,} leg, head) of human instances, resulting in confusing prediction of the unlabeled keypoints, as shown in Figure~\ref{fig:limitaion}. We will conduct deep studies on this problem in future works.

\end{appendices}

\end{document}